\documentclass[conference]{IEEEtran}
\IEEEoverridecommandlockouts
\usepackage{cite}
\usepackage{amsmath,amssymb,amsfonts}
\usepackage{hyperref}
\usepackage{algorithmic}
\usepackage{graphicx}
\usepackage{textcomp}
\usepackage{xcolor}
\def\BibTeX{{\rm B\kern-.05em{\sc i\kern-.025em b}\kern-.08em
    T\kern-.1667em\lower.7ex\hbox{E}\kern-.125emX}}
\begin{document}

\title{An Overview of Deep Learning Architectures in Few-Shot Learning Domain
}

\author{\IEEEauthorblockN{Shruti Jadon\thanks{This work has been as part of writing "Hands on one-shot Learning using Python" book.}}
\IEEEauthorblockA{\textit{IEEE Member} \\
shrutijadon@ieee.org}
\and
\IEEEauthorblockN{Aryan Jadon}
\IEEEauthorblockA{\textit{San Jose State University}\\
aryan.jadon@sjsu.edu}
}

\maketitle

\begin{abstract}
   Since 2012, Deep learning has revolutionized Artificial Intelligence and has achieved state-of-the-art outcomes in different domains, ranging from Image Classification to Speech Generation. Though it has many potentials, our current architectures come with the pre-requisite of large amounts of data. Few-Shot Learning (also known as one-shot learning) is a sub-field of machine learning that aims to create such models that can learn the desired objective with less data, similar to how humans learn. In this paper, we have reviewed some of the well-known deep learning-based approaches towards few-shot learning. We have discussed the recent achievements, challenges, and possibilities of improvement of few-shot learning based deep learning architectures. Our aim for this paper is threefold: (i) Give a brief introduction to deep learning architectures for few-shot learning with pointers to core references. (ii)
Indicate how deep learning has been applied to the low-data regime, from data preparation to model training. and, (iii) Provide a starting point for people interested in experimenting and perhaps contributing to the field of few-shot learning by pointing out some useful resources and open-source code.
Our code is available at Github:\url{https://github.com/shruti-jadon/Hands-on-One-Shot-Learning}

\end{abstract}
\section{Introduction}
Humans learn new things with a very small set of examples e.g. a child can generalize the concept of a ”Dog” from a single picture but a machine learning system needs a lot of examples to learn its features. In particular, when presented with stimuli, people seem to be able to understand new concepts quickly and then recognize variations on these concepts in future precepts. Machine learning as a field has been highly successful at a variety of tasks such as classification, web search, image and speech recognition. Often times however, these models do not do very well in the regime of low data. This is the primary motivation behind Few Shot Learning; to train a model with fewer examples but generalize to unfamiliar categories without extensive retraining. Deep learning has played an important role in the advancement of machine learning, but it also requires large data-sets. Different techniques such as regularization reduces over-fitting in low data regimes, but do not solve the inherent problem that comes with fewer training examples. Furthermore, the large size of data-sets leads to slow learning, requiring many weight updates using stochastic gradient descent. This is mostly due to the parametric aspect of the model, in which training examples need to be slowly learned by the model into its parameters. In contrast, many known non-parametric models like nearest neighbors do not require any training but performance depends on a sometimes arbitrarily chosen distance metric like the L2 distance.few-shot learning is an object categorization problem in computer vision. Whereas most machine learning based object categorization algorithms require training on hundreds or thousands of images and very large data-sets, few-shot learning aims to learn information about object categories from one, or only a few, training images \cite{jadon2020hands}.
\begin{figure}[!ht]
\centering
  \includegraphics[scale=0.25]{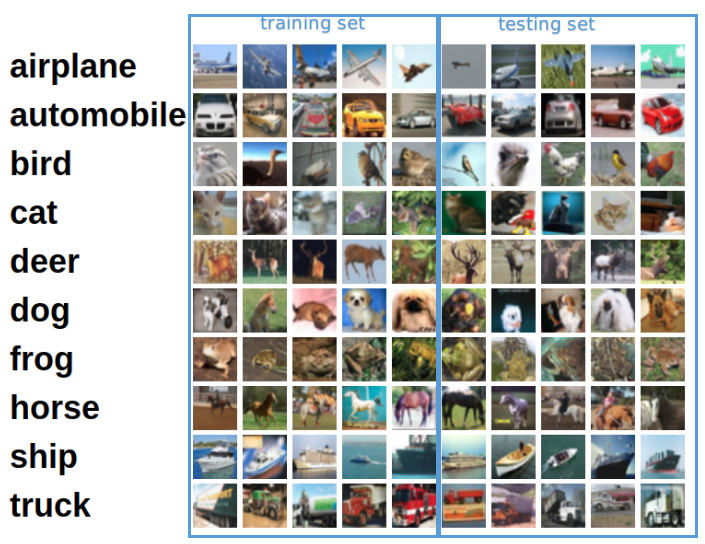}
  \includegraphics[scale=0.15]{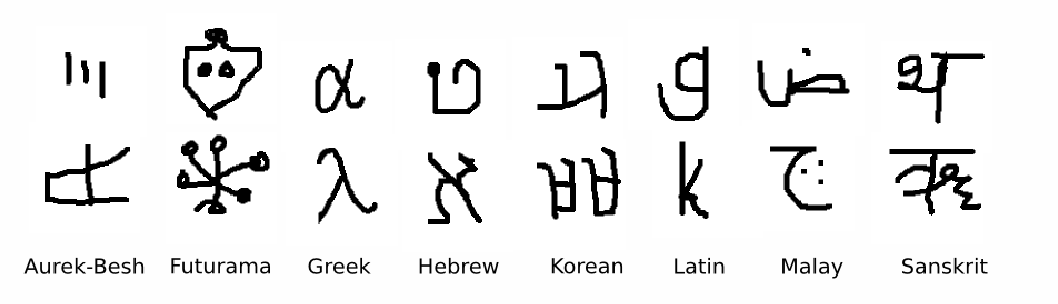}
  \caption{Sample images of widely used few-Shot learning data-sets:mini-Imagenet and Omniglot data-set}
  \label{fig:1}
\end{figure}
\subsection{Data-sets for Few-shot learning}
Most commonly used data-sets for few shots learning are Mini-ImageNet, and Omniglot data-set. Omniglot is a data-set is specially designed to compare and contrast
the learning abilities of humans and machines. The data-set contains handwritten characters of 50 languages (alphabets) with 1623 total characters. The data-set is divided into a background set and an evaluation set. Background set contains 30 alphabets (964 characters) and only this set should
be used to perform all learning (e.g. hyper-parameter inference or feature learning). The remaining 20 alphabets are for pure evaluation purposes only. Each character is a 105 x 105 greyscale image.
There are only 20 samples for each character, each drawn by a distinct individual(Refer Figure 2). \\
Mini-Imagenet as the name suggests if the small version of Image-Net data-set, specifically designed for Few-Shots Learning Image Classification Task, Refer Figure \ref{fig:1}.

\section{Types of Few-Shot Learning Approaches}
In recent years, Deep Learning field has advanced a lot, It grew from a simple Artificial Neural Network classification to LSTM Networks for Language Modeling and Convolutional Neural Networks for Image processing. Though Deep Learning has provided many solutions for various problems such as face recognition, speech recognition etc; but there are still major industries, which faces the constraints of smaller data-set are yet to experience all the advantages of Deep Learning, such as the medical and manufacturing industries. A lot of research has been done in figuring out architectures to learn from less data regime. In this paper we have explained some widely used few-shot deep learning architectures and attempted to explain the intuition and logic behind these architectures. In broader sense, deep learning based few-shot learning approaches majorly can be divided in 4 categories: 

 \begin{itemize}
\item Data Augmentation Methods.
\item Metrics Based Methods. 
\item  Models Based Methods.
\item  Optimization Based Methods.

\end{itemize}

\section{Data Augmentation Methods}
Data Augmentation is a technique to enhance the amount and enrich the quality of training data-sets. It has been majorly appreciated by vision deep learning community to build better models. Image Augmentation methods have evolved in recent years from simple geometric transformations, color space augmentations to generative adversarial networks, neural style transfer methods. At present, a lot of research is being conducted in application of augmentation methods based on GANs for their practical use-case in medical field community \cite{jadon2020comparative}. In addition to general augmentation techniques, use of Data Augmentation can also improve the performance of models and expand limited data-sets to take advantage of the capabilities of big data. Though Data Augmentation is a good approach to solve less data problem, but it also have some limitations, for example, If your current data distribution is skewed, that will result in augmented data distribution to be skewed as well. There is also a high chance of over-fitting with data augmentation.

\section{Metrics Based Methods}
Metrics Based Methods, as the name suggests is based upon metrics. Metrics play a very important role in our models, as we know the underlying idea of all machine learning or deep learning models is mathematics, we convey our data and objective to machines through the mathematical representation of data, but this representation \cite{jadon_2018} can have many forms especially if we go in higher dimensions, or if we wish to learn these representations based on our different objectives; for example, to calculate similarity between 2 images, we can calculate both Euclidean distance or cosine similarity. The key question is how to learn which embeddings are better representation of our task, or which loss function is good for our objective. In this section, we have summarized two well known metrics based approaches: Siamese Networks and Matching Networks, which attempts to improve the task representative embedding through different architecture and training procedure.
\begin{figure}[t]
\begin{center}
  \includegraphics[width=0.7\linewidth]{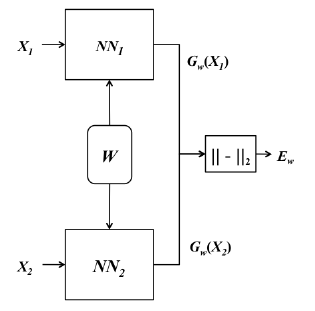}
\end{center}
  \caption{Sample Siamese Network Architecture}
\label{fig:3}
\end{figure}

\subsection{Siamese Networks}
A Siamese network\cite{koch2015siamese}, as the name suggests, is an architecture with two parallel layers. In this architecture, instead of a model learning to classify its inputs using classification loss functions, the model learns to differentiate between two given inputs. It compares two inputs based on a similarity metric, and checks whether they are same or not. This network consists of two identical neural networks, which share similar parameters, each head taking one input data point (see fig. \ref{fig:3}). In the middle layer, we extract similar kinds of features, as weights and biases are the same. The last layers of these networks are fed to a loss function layer, which calculates the similarity between the two inputs. The whole idea of using Siamese architecture\cite{jadon2019improving} \cite{jadon2020hands} is not to classify between classes but to learn to discriminate between inputs. So, it needed a differentiating form of loss function such as the contrastive loss function. 
\begin{figure*}[!ht]
\begin{center}
\includegraphics[scale=0.40]{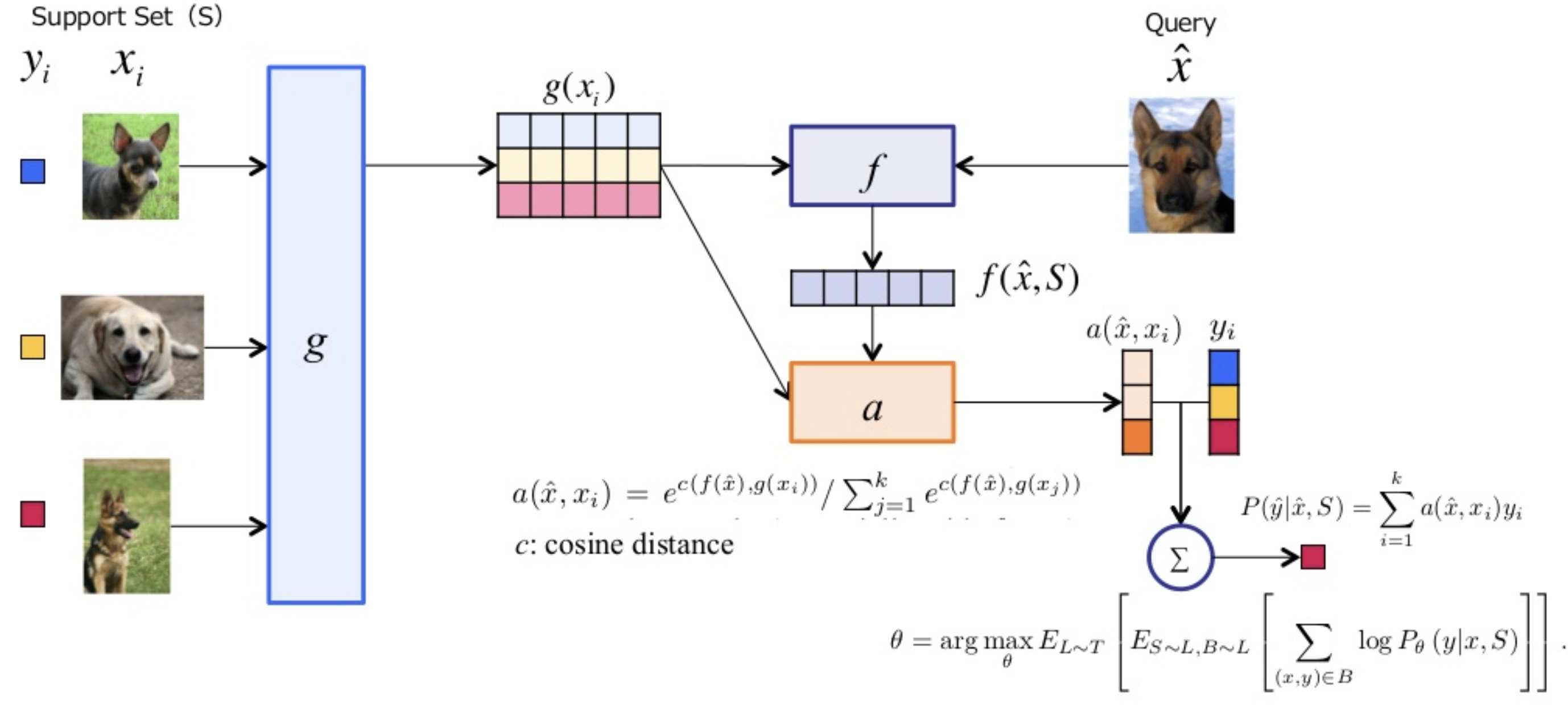}
\end{center}
  \caption{In the following sample architecture for matching networks, input to g($\theta$) are randomly sampled data (support set) from already existing examples, and input to f($\theta$) is input test data point}
\label{fig:short}
\end{figure*}
\subsubsection{Understanding the data processing for Siamese Networks} is really important. For training Siamese network, we need a special kind of pre-processing the data-set. We have to create pairs of data points:
\begin{itemize}
\item pair of similar images, and
\item pair of dissimilar images.
\end{itemize}
We also need to create labels accordingly for similar($y=1$), and dissimilar data points($y=0$); then, each pair is fed to the Siamese architecture. At the end of the layer, Siamese Network uses a differentiating form of loss function known as Contrastive Loss Function.
\subsubsection{Understanding Contrastive loss function}
As the whole idea of using Siamese architecture is not to classify between classes but to learn to discriminate between inputs. So, it needed a differentiating form of loss function known as contrastive loss function.  It is given as follows:
$$ Loss=(1-Y)\frac{1}{2}D_w^2+(Y)\frac{1}{2}(max(0,m-D_w)^2 $$
\\
Here, $D_w=\sqrt{{G_w(X1)-G_w(X2}}$ and $G_w$ is the output from one of sister networks, $X1,X2$ are input data pair. 

If we carefully look into the equation of Contrastive Loss Function, we will observe that it consists of dual terms:
\begin{itemize}
    \item first part to decrease the energy of like pairs and,
    \item second increase the energy of unlike pairs.
\end{itemize}
The first part of the loss function is simply Mean Squared Error multiplied with their respective labels; whereas, the second part of loss function resembles Hinge Loss, with m as a threshold. Here $m$ is a margin value which is greater than 0. It indicates that dissimilar pairs with Root Mean Squared Value greater than $m$, won't contribute in the dissimilar pair loss function. This also holds up logically as you would only want to optimize the network based on pairs that are actually dissimilar, which network thinks are similar. 

Many times a problem can be solved using various approaches, for example, Face Detection in our phones, Signature verification, etc. Image Classification is one of the approaches but that will need a lot of data points; whereas, if we use the Siamese network architecture of One-shot learning, we can achieve greater accuracy with only a few data points. Siamese Network Architecture became one of the popular One-shot learning architecture, which industry has adopted and are using for various other applications, such as Face Detection, Handwriting Detection, Spam Detection, and so on. But there is still a lot of scope of improvement, for example, for Contrastive Loss Function itself. In other modfied versions of Siamese Networks Architecture,  it was discussed that contrastive loss functions, wasn't able to learn through decision boundaries very clearly, so a new loss function was suggested known as Triplet Loss, which helped the architecture to advance.

\textbf{Note:} Transfer learning using Siamese Network Approach is only valid for similar domain dataset \cite{vargas2020one}. If the domain aren't similar, we also needs to take care of domain adaption, that is, try to ensure that our training and testing data-sets are close in terms of data distribution. For example, if one wants to create a system to test whether two hand writings are of the same person, what they can do is train Siamese Network Architecture on MNIST data-set through which it will learn handwriting specific features such as curves, strokes of any character.
\subsection{Matching Networks}
Matching networks \cite{vinyals2016matching}, in general, proposes a framework which learns a network that maps a small data-set and an unlabeled example to its label. When it comes to a small data-set, we come across the problem of over-fitting and under-fitting which can be alleviated using regularization and data augmentation respectively; but still, it doesn't solve the root cause of the issue. The other problem is that learning of parameters is very slow, requiring various weight updates using stochastic gradient descent; whereas many non-parametric models, such as k-nearest neighbor, doesn't even require any form of training; it stores data-set and makes a decision based on a metric. Matching networks incorporate the best characteristic of both parametric and non-parametric models, also famously known as Differential Nearest neighbor.
Matching networks are designed to be two-fold:
\begin{enumerate}
    \item \textbf{Modeling level:} At Modeling level, they proposed Matching nets, which uses advances made in attention and memory that enable fast and efficient learning.
    \item \textbf{Training procedure:} At Training Level, they have one condition that distribution of training and test set must be same. For example: show a few examples per class, switching the task from mini-batch to mini-batch, similar to how it will be tested when presented with a few examples of a new task.
\end{enumerate}

In simpler terms, matching networks learn the proper embeddings representation and use cosine similarity measure to ensure whether a test data point is something ever seen or not. 

Matching Networks is a neural network model that implements an end-to-end training procedure that combines feature extraction and differentiable k-NN with cosine similarity.
Matching network's architecture is majorly inspired by the attention model and memory-based networks. In all these models, a neural attention mechanism is defined to access a memory matrix, which stores useful information to solve the task at hand. To begin with, first we need to understand certain terminologies of Matching Network:
\begin{enumerate}
    \item \textbf{Label Set:} This is the sample set of all possible categories. For example, if we have 1,000 categories and as part of sampling from the database we only use only 5 categories.
    \item \textbf{Support Set:} This is the sampled input data points (for example, images) of label set categories.
    \item \textbf{Batch:} Similar to support set, batch is also a sampled set consisting of input data points of Label Set Categories.
    \item \textbf{N-way K-Shot Method:} Here, N is the size of the support set. In simpler terms, the number of possible categories in the training set. For example, in the figure that follows, we have four different types of dog breeds, and we are planning to do 5-shot learning method, that is, have at least five examples of each category. This will make Matching Network Architecture as 4 way 5 shot learning.
\end{enumerate}

The key idea of Matching networks is to map images to an embeddings space, which also encapsulates the label distribution and then project test image in the same embedding space using different architecture; later, use cosine similarity to measure the similarity metric. 
\subsubsection{Understanding data processing for Matching Networks}
For Matching networks architecture, we need to sample N labels from training dataset, and then according to those labels, sample K examples for sample-set, and B examples for batch-set. As part of pre-processing data, a support set S of k examples will be created as $(x_i,y_i)$ as shown in fig \ref{fig:3}. After obtaining Support Set, it passes through a standard feature extraction layer ($g$), followed by bidirectional LSTM architecture, it helps us learn the probabilistic distribution of labels present in support set, by switching from batch to batch. These extracted embeddings are called Fully Contextual Embeddings.

Similarly, for the query image $ \hat{x}$ also go through similar path and obtain Full Context Embeddings in same embedding space as that of Support set. After obtaining the outcomes from both support set and query image, they get feeded into a kernel followed by our cosine similarity function.
\subsubsection{Understanding Training Process of Matching Networks} 
Training process of matching networks is a bit different from traditional deep learning approaches. It tries to replicate data distribution of test-data in training data as well. In simpler terms, as we have observed as part of data pre-processing step, Matching Networks actually create sample sets from training examples too and try to re-create scenario similar to test-sets. Matching Networks learns its parameters through training-set by creating small subsets \cite{jadon2020ssm} as that of a test-sets; that is, it is actually training the model as a few-shot learning model.

\section{Models Based Methods}
Models Based Methods are majorly inspired from how humans store prior information in memory units, and have access while learning new objectives.
Model-Based Meta-Learning models depend upon architecture design for rapid generalization of few-shot learning tasks. As Humans, we always have access to a large amount of memory for any task, and a majority of the times, how humans learn a new task is through recalling some memory and adapt to the new task in short span. In this chapter, we will be studying about such deep learning architectures, which are able to learn parameters with only a few training steps with the help of external memory, such for of architectures are known as Model-Based methods.

\begin{figure}[!ht]
\begin{center}
\includegraphics[scale=0.45]{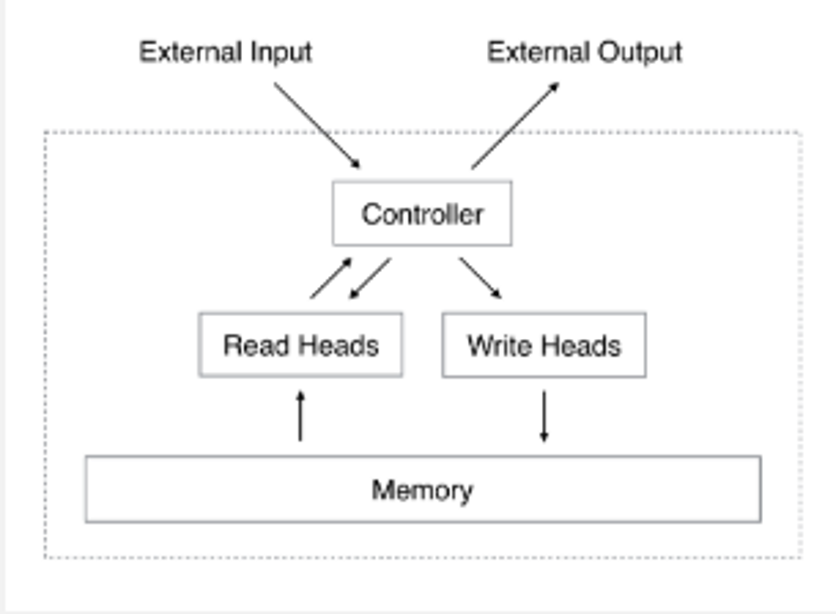}
\end{center}
  \caption{Neural Turing Machine\cite{graves2014neural}}
\label{fig:4}
\end{figure}
\begin{figure*}[!ht]
\begin{center}
\includegraphics[scale=0.45]{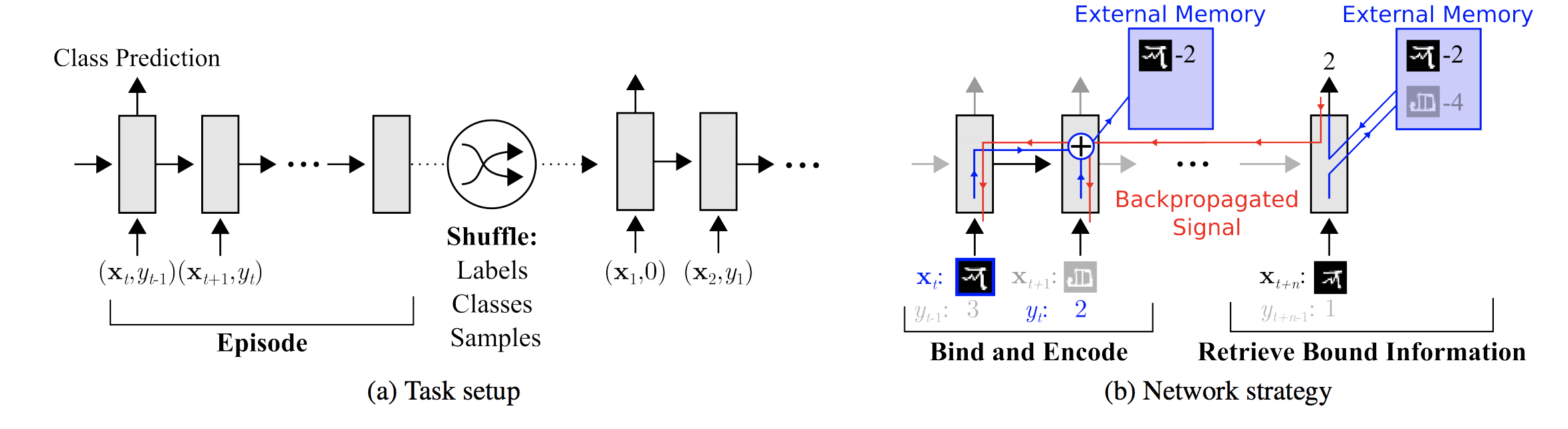}
\end{center}
  \caption{Memory Augmented Neural Network Approach}
\label{fig:short}
\end{figure*}
\subsection{Neural Turing Machine}
The early development stages of the field of Artificial Intelligence was heavily dominated with the symbolic approach - which attempted to explain how information processing systems like human brain might work in terms of symbols, structures and rules to manipulate those symbols and structures. It wasn't until about 1980s when the field of AI took a different approach - connectionism, neural networks being the most promising modeling technique of connectionism. The advent of neural networks met with two heavy criticisms. First neural networks typically accept fixed sized inputs and won't be much useful as most of the input in real life problems is variable length (sequence of frames in video, text, speech). Second, neural networks seem unable to bind values to specific locations within data structures which is heavily employed by two information systems we know of - human brain and computers. The first criticism about variable length input was answered by Recurrent Neural Networks (RNNs) which have achieved state-of-the-art performance on many tasks. Neural Turing Machines \cite{graves2014neural}, the topic of discussion in this section tries to answer second criticism by giving external memory to neural networks and a method to learn to use it. We will discuss the overall architecture of Neural Turing Machine (NTM) in this section which is foundational to understanding Memory-Augmented Neural Networks (MANN) \cite{santoro2016one} which modifies NMT architecture and adapt it for few-shot learning task.

Modern computers have advanced a lot over past 50 years but they are still composed of three systems - memory, control flow and arithmetic/logic operations.  There's also biological evidence that working memory is crucial in quick and meaningful store and retrieval of information. The same is also established extensively by the computational neuroscience field. Taking inspirations from this, A NTM \cite{graves2014neural} is fundamentally composed of a neural network - controller and a 2D matrix called the memory bank (or memory matrix). At each time step, the neural network receives some input and generates some output corresponding to that input. In the process of doing so, it also accesses the internal memory bank and performs read and/or write operations on it. Drawing inspiration from traditional turing machines, NMT uses the term 'head' to specify the memory location(s). The overall architecture is shown in Figure \ref{fig:4}. There's one challenge in this architecture, If we access the memory location by specifying row and column index in the memory matrix, we can't take gradient of that index. This operation is not back-propagable and would restrict the training of NMT using standard back-propagation and gradient descent based optimization techniques. To circumvent this problem, the controller of the NTM \cite{collier2018implementing} interacts with memory using 'blurry' read and write operations that interacts with all elements of the memory to varying degrees. More precisely the controller produces weights over all memory locations in a differentiable manner which helps in training the network end-to-end using standard gradient based optimization methods.
\begin{enumerate}
    \item \textbf{Modeling}: The memory matrix at time step $t (M_t)$ has $R$ rows and $C$ columns. There's an attention mechanism which dictates where the head should read/write from. This attention vector, generated by controller is a length- $R$ vector called weight vector $W_t$ where each entry of this vector $W_t(i)$ is the weight for the $i^{th}$ row of the memory bank. The weight vector is normalized which means it satisfies following conditions:
    
    $0 \leq w_t(i) \leq 1$ and , $\sum_{i=1}{R}w_t(i)=1$
    
    \item \textbf{Reading}: The read head will return a length- $C$ vector $r_t$ that is a linear combination of the memory's rows $M_t(i)$ scaled by the weight vector:
    
    $r_t \gets \sum_{i=1}{R}(w_t(i)M_t(i))$
    \item \textbf{Writing}: Writing is a combination of two steps: erasing and adding. In order to erase old data the write head uses an additional length- $C$ erase vector, $e_t$   along with the weight vector. The following equations defines the intermediate step of erasing the rows. 
    
    $M_t^{erased}(i) \gets M_{t-1}(i)[1-w_t(i)e_t)$
    Finally, the write head uses length- $C$  add vector $a_t$  along with $M^{erased}$ from above and weight vector to update the rows of the memory matrix.
    $M_t(i) \gets M_t^{erased}+w_t(i)a_t$
    \item \textbf{Addressing}: The key to read and write operation is the weight vector which indicates which rows to read/write from/to. The controller produces this weight vector in four stages. Each stage produces in an intermediate vector which gets passed to the next stage. The first stage is content-based addressing, the goal of which is to generate a weight vector based on how similar each row is to some key vector $k_t$ of length $C$. More precisely, controller emits vector $k_t$  which is compared to each row of $M_t$ using cosine similarity measure, defined below
    
    $K(u,v)=\frac{u.v}{||u||.||v||}$
    
    The content weight vector is not normalized yet, so its normalized with the following operation.
    
    $w_t^c(i)=\frac{exp(\beta_tK(k_t,M_t(i)))}{\sum_j\exp(\beta_t K(k_t, M_t(j)))}$
    
    The next stage is the location based addressing which focuses on read/write from specific memory locations as opposed to read/write specific location values done by stage 1. In the second stage, a scalar parameter $g_t \in(0,1)$ called the interpolation gate, blends the content weight vector $w_t^c$  with the previous time step's weight vector $w_{t-1}$ to produce the gated weighting $w_t^g$ . This allows the system learn when to use (or ignore) content-based addressing.
    
    $w_t^g \gets g_tw_t^c+(1-g_t)w_{t-1}$
    
    In the third stage, after interpolation, the head emits a normalized shift weighting $s_t$   to perform shift modulo $R$  operation (i.e. move rows upwards or downwards). This is defined by the operation below.
    
    $\tilde{w_t}(i) \gets \sum_{j=0}^{R-1}w_t^g(j)s_t(i-j)$
    
    The fourth and final stage, sharpening, is used to prevent the shifted weight $\tilde{w_t}$ from blurring. This is done with a scalar $\gamma \geq 1$ and applying the operation.
    
    $w_t(i) \gets \frac{\tilde{w_t}(i)^{\gamma_t}}{\sum_{j}\tilde{w_t}(j)^{\gamma_t}}$
\end{enumerate}
All the operations, including read, write and four stages of addressing are differentiable and thus the entire NMT model could be trained end-to-end with back-propagation and any gradient descent based optimizer. The controller is a neural network which could be a feed-forward network or even a recurrent neural network like LSTMs.
Now that we understand the architecture and the working of NTM, we can dive into Memory Augmented Neural Networks which is a modification of NMT and has been modified to excel at few-shot learning. 

\subsection{Memory Augmented Neural Networks}
The goal of Memory Augmented Neural Network (MANN) \cite{santoro2016one} is to excel at few-shot learning task. The NMT controller, as we read earlier uses both content-based addressing and location-based addressing. On the other hand, the MANN controller uses only content-based addressing. There are two reasons to do this. One reason is that location-based addressing is not required for the few-shot learning task. In this task, for a given input, there are only two actions a controller might need to take and both actions are content dependent and not location dependent. One action is that the input is very similar to previously seen input in which case we might what to update whatever is there in memory. The other action is that current input is not similar to previously seen inputs in which case we don't want to overwrite the recent information but the least used memory location. The memory module, in this case, is called Least Recently Used Access (LRUA) module.
\begin{enumerate}
    \item \textbf{Reading:} The read operation of MANN is very similar to the read operation of NTMs, with a minor difference that the weight vector here uses only content bases addressing (stage -1 of NMT addressing). More precisely, the controller uses normalized read weight vector, $w_t^r$ is used along with the rows of the $M_t$  to produce read vector $r_t$.
    $r_t \gets \sum_{i}^{R}(w_t^r(i)M_t(i)$
    The read-weight vector $w_t^r$ is produced by controller defined by the operations below. 
    $w_t^r=\frac{exp(K(k_t,M_t(i))}{\sum_j exp((k_t,M_t(j)))}$
    where, operation $K()$ is the cosine similarity , similar to the one defined for NMTs above. 
    \item \textbf{Writing:} To write to the memory, the controller interpolates between writing to the most recently read memory rows and writing to least recently read memory rows.
    $w_t^w \gets \sigma(\alpha)w_{t-1}^r+(1-\sigma(\alpha))w_{t-1}^{lu}$
    $M_t(i) \gets M_{t-1}(i) + w_t^w(i)k_t$
    $w_t^u \gets \gamma w_{t-1}^u + w_t^r + w_t^w$
\end{enumerate}

\subsection{Meta Networks}
Meta Networks \cite{munkhdalai2017meta} as the name suggests is a form of Model-based Meta-Learning Approach. Similar to how Meta-Learning approaches, In meta-networks also, we have a base learner and meta-learner sharing parameters where a meta-learner extracts common features of all tasks and a base learner learns the targeted task. As we can see in figure \ref{fig:6}, Meta Networks is also equipped with external memory, similar to architecture we just read above Memory Augmented Neural Networks. The underlying key idea of Meta Networks is to learn weights in a fast manner for rapid generalizations, and it has been done by processing higher order meta information. In the next section, we will go through an understanding of architecture, extraction of meta information, and it's parameter optimization algorithm. 

\textbf{Note:} Previous work on Meta-learning has formulated the problem as two-level learning: slow learning of a meta-level model performing across tasks and rapid learning of a base-level model acting within each task. The goal of a meta-level learner is to acquire a general knowledge of different tasks. The knowledge can then be transferred to the base-level learner to provide generalization in the context of a single task. The base and meta-level models can be framed in a single learner.

In usual Deep Learning Methods, weights of neural networks are updated by stochastic gradient descent and many times it takes a lot of time to train, in other words, slow weights update. In Meta Networks, one solution has been suggested to improve upon this method: To utilize one neural network in parallel to original neural network, to predict the parameters of another neural network and the generated weights are called fast weights.

In MetaNet \cite{munkhdalai2017meta}, loss gradients are used as meta information to enable models that learn fast weights. Slow and fast weights are combined to make predictions in neural networks just as shown in the figure above. Here, $sum$ means element-wise sum.
Meta Networks Learning occurs at two levels in separate spaces: meta space and task space. here, the base learner learns in task-specific space, whereas the meta-learner learns meta-knowledge across different tasks. After training of Meta learner, the base learner first analyzes the input task and then provides the meta-learner with feedback in the form of higher order meta information(gradients) to explain its own status in the current task space.
\begin{figure}[t]
\begin{center}
   \includegraphics[width=1.0\linewidth]{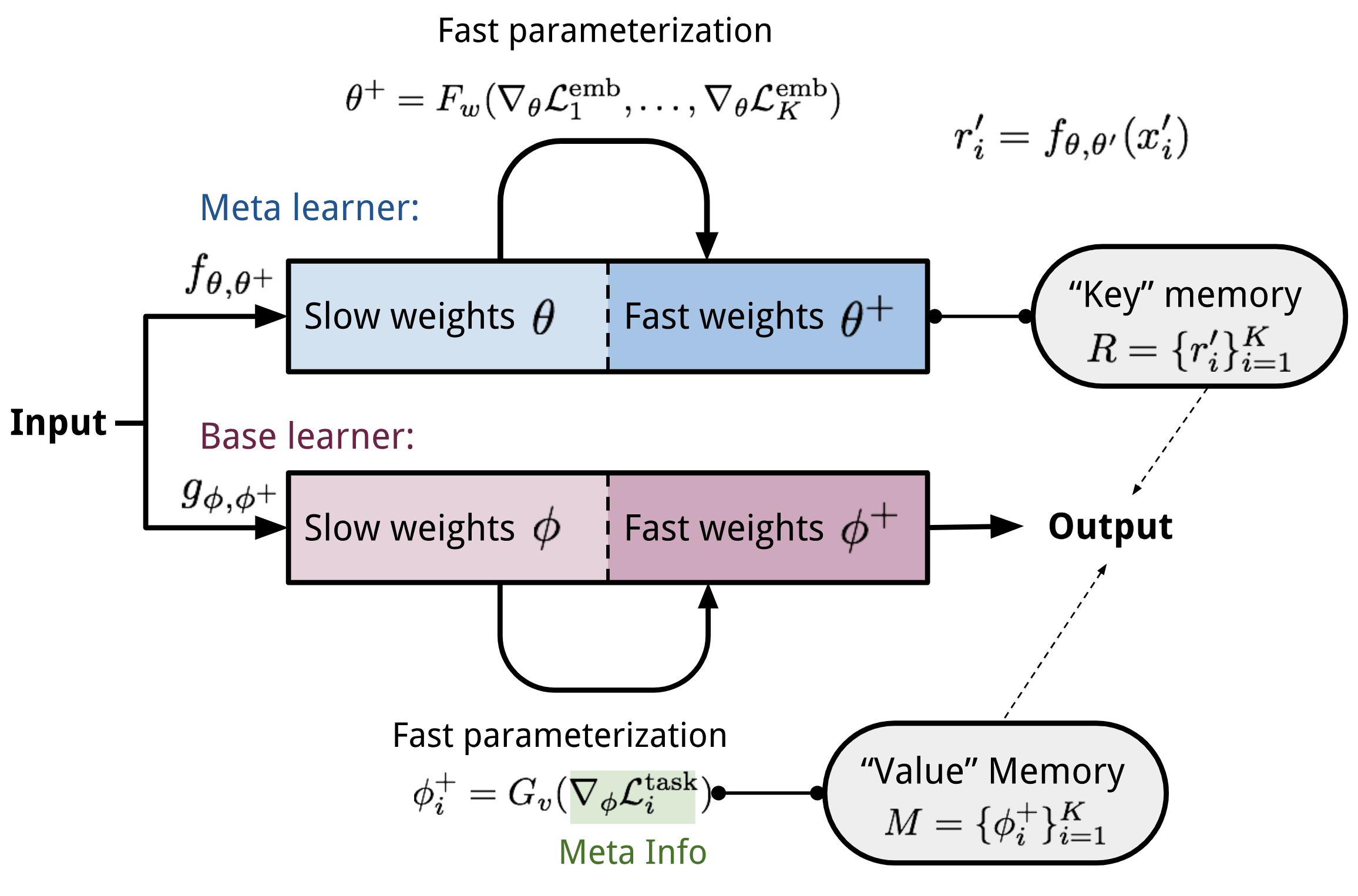}
\end{center}
   \caption{Architecture of Meta Networks}
\label{fig:7}
\end{figure}

There are two important components of a MetaNet as shown in figure \ref{fig:7}:
\begin{enumerate}
    \item \textbf{Embedding function $(f_\theta)$:} As part of Meta Net, we train Embedding generator function, which is then used to compare features of two different data points, very similar to what we have in Siamese Networks.
    \item \textbf{Base Learner Model $g_\phi$:} It is parameterized by weights, and completes the actual learning task.
\end{enumerate}

As we have two different types of networks trained in parallel: fast, and slow network, therefore we also need to get Embedding Function and a Base Learner Model for a fast network:
\begin{enumerate}
    \item Embedding Function$(F_w)$: an LSTM architecture, for learning fast weights $\theta$ of the embedding function$(f_\theta)$ of slow network.
    \item Base Learner Model $(G_v)$: a neural network parameterized by v learning fast weights $\phi$ for the base learner $g_\phi$from its loss gradients.
\end{enumerate}
For training, Meta Networks also follow a similar training procedure as of Matching Networks. Here, also we obtain training data contains multiple pairs of datasets: support set S = $(x_i',y_i')$ and training set U= $(x_i,y_i)$ . So, now we have four networks and four sets of model parameters to learn, $(\theta,\phi,w,v)$. \cite{munkhdalai2017meta} also proposed the meta learning based algorithm to train networks as shown in figure \ref{fig:6}

\begin{figure}[t]
\begin{center}
   \includegraphics[width=1.0\linewidth]{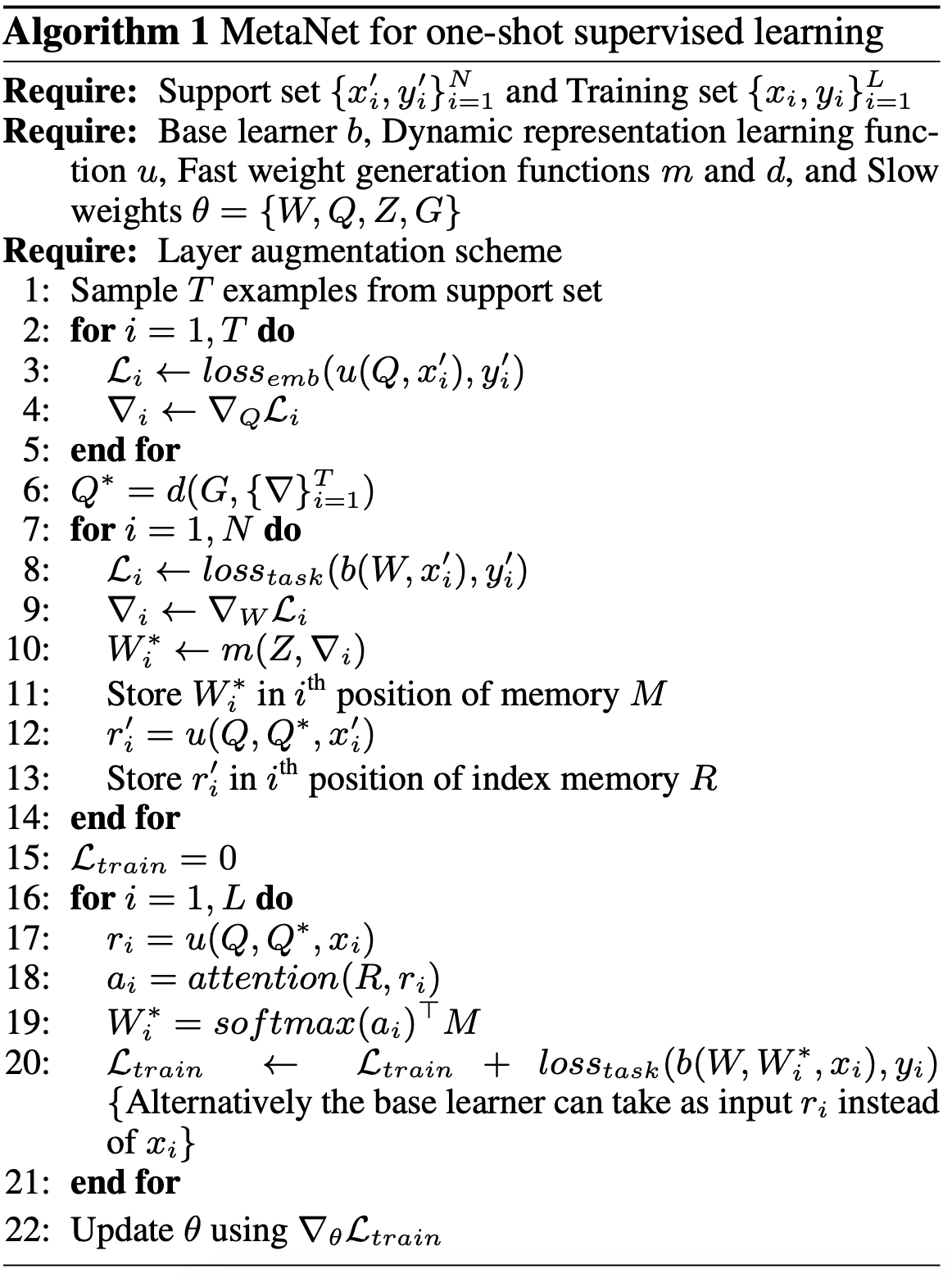}
\end{center}
   \caption{Algorithm for One-shot learning using Meta-Net in Supervised data setting \cite{munkhdalai2017meta}.}
\label{fig:6}
\end{figure}

\section{Optimization Based Methods}
If we look into the learning method of Neural Networks architectures, it usually consists of a lot of parameters and is optimized using Gradient descent algorithm, which takes many iterative steps over many examples to perform well. Gradient Descent Algorithm though provides a decent performance of models, but fails drastically in scenarios when there is less amount of data. There are mainly two reasons why gradient descent algorithm fails to optimize a neural network with smaller data-set:
\begin{enumerate}
    \item If we look into variants of gradient descent's weight updating step methods (Adagrad, Adam, RMS, and so on), can't perform well with less number of epochs. Especially when used for non-convex optimization these algorithms don't have a strong guarantee of convergence.
    \item For each new task, Neural Network has to start from a random initialization of its parameters, which results in late convergence. Transfer learning has been used to alleviate this problem, by using a pre-trained network, but it is limited to be of same domain task.
\end{enumerate}
What can be really helpful is to learn some common initialization, which can be used across all domains as a good point of initialization. The key idea of a Gradient Descent Algorithm is based on the direction of the next step, which is chosen on the basis of probabilistic distribution assumption. So, if we somehow are able to approximate that probabilistic distribution completely, we will be able to optimize the network, with only a few steps. This is the basic idea of Optimization-based algorithms: Model-Agnostic Meta-Learning \cite{finn2017model} and LSTM-Meta Learner \cite{ravi2016optimization}
\begin{figure}[t]
\begin{center}
   \includegraphics[width=1.3\linewidth]{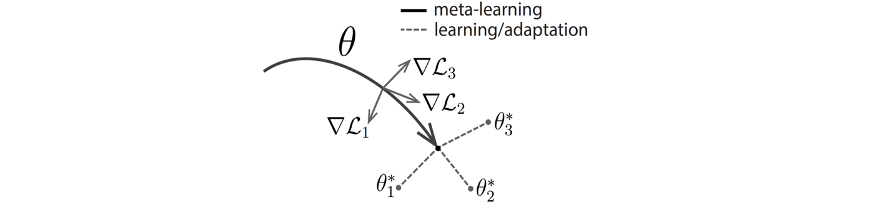}
\end{center}
   \caption{Here, $\theta$ is the model’s \cite{finn2017model} parameters and the bold black line is the meta-learning phase. Let's assume that we have 3 different new tasks 1, 2 and 3, a gradient step is taken for each task (the gray lines)}
\label{fig:8}
\end{figure}
 \begin{figure}[t]
\begin{center}

  \includegraphics[width=1.0\linewidth]{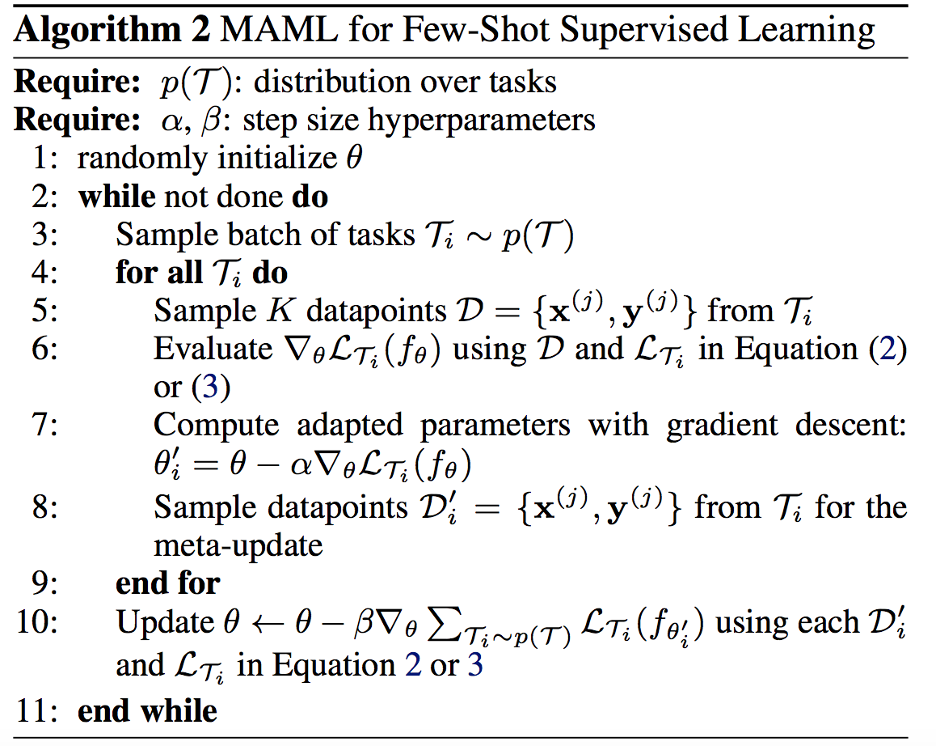}
\end{center}
  \caption{Optimization Algorithm for Model Agnostic Meta Learning \cite{finn2017model}.}
\label{fig:9}
\end{figure}

\subsection{Model Agnostic Meta Learning}
 The goal of Model Agnostic Meta-Learning \cite{finn2017model} architecture is to quickly learn a new task from a small amount of new data. The key idea of this approach is to train the models' initial parameters using a different data-set, such that as the new task comes, the model gives maximum performance by using already initialized parameters, to fine-tune the architecture through one or more gradient steps. As we know that Neural Networks use Gradient descent algorithm to train, so the method of training a model's parameters such that a few gradient steps can optimize loss function, and give good results, can be viewed from a feature learning standpoint as building an internal representation. As it is very generic, therefore can be used for various tasks. The primary contribution of this work is a simple model and task-agnostic algorithm that trains a model's parameters such that a small number of gradient updates will lead to fast learning on the new task. The Objective of Model-Agnostic Meta-Learning (MAML) \cite{finn2017model} is to provide a good initialization of a model’s parameters to achieve optimal fast learning on a new task with less number of gradient steps. It also attempts to avoid overfitting scenarios, which happens while training a Neural Network with less amount of data architecture. 

As we can see in the figure \ref{fig:8} parameters $\theta$ are close to all the 3 optimal parameters of task 1, 2, and 3 which makes $\theta$ the best parameters initialization that can quickly adapt to different new tasks. As a result, only a small change in the parameters $\theta$ will lead to an optimal minimization of the loss function of any task. This is the key idea of MAML, that somehow we learn $\theta$ through the original dataset, so while fine-tuning on the real dataset, we just have to move a small step.
For One/Few Shots Learning, first, we prepare our data. For a particular task say, $T_i$  we have data samples which are forming task distribution of form $P(T)$ (Probabilistic distribution across all tasks). Our aim is to train a model to learn task  $T_i$ from K data points and feedback generated by  Loss , $L(T_i)$.  During meta-training, a task $T_i$ is sampled from $P(T)$ i.e; training data available, then model is trained with K samples and feedback from the corresponding loss $L(T_i)$ from $T_i$ , and then tested on new samples from $T_i$ . The model is then improved by considering how the test error on new data changes with respect to the parameters. In effect, the test error on sampled tasks $T_i$ serves as the training error of the meta-learning process. At the end of the training, new tasks are sampled from $p(T)$, and meta-performance is measured by the model’s performance after learning from K samples.

$L_{T_i}(f_{\phi})=\sum_{x^{j},y^{j} \backsim T_i}||f_{\phi}(x^{(j)})-y^{(j)}||_{2}^{2}$

where $x^{(j)},y^{(j)}$ are sampled from task $T_i$. To learn more in detail about steps followed in maml optimization Algorithms, please take a look at the algorithm published in the paper \cite{finn2017model}.
 
\subsection{LSTM Meta Learner}
As the name suggests, LSTM meta-learner \cite{ravi2016optimization} algorithm is specifically designed on LSTM Architecture because a cell-state update in LSTM is similar to gradient-based update in backpropagation, and information about the history of gradients help in optimizing model better. This Optimization Algorithm is trained to optimize a learner neural network classifier. It captures the knowledge of both the short term of a particular task and common long term.

LSTM meta learner is heavily inspired from connection in logic of Gradient descent and LSTM cell update. Ravi et al., \cite{ravi2016optimization} explained that iff we look into update method of Gradient Descent, we will see an equation like this:

$\theta_t = \theta_{t-1}-\alpha_t \bigtriangledown L_t$

here, $\theta_t$ is parameters at t time step, $\bigtriangledown L_t$ is gradient of Loss at t, and $\alpha_t$ is learning rate at time t.

One observation that authors of LSTM meta learner made was, that this update looks very similar to how cells get updated in LSTMs.. So, The Key idea of Meta-Learner LSTM is to learn an update rule for training a neural network. Lets first dive in how a cell state is defined in an LSTM.

$c_t = f_t\odot c_{t-1} +i_t\odot \bar{c_t}$

If we put $f_t=1, c_{t-1}=\theta_{t-1}, i_t=\alpha_t$ and $\bar{c_t}=\bigtriangledown L_t$ , we will get a gradient descent update rule.
Considering this, Logically, we want to learn $i_t$, as that is essentially similar to estimating the learning rate of gradient descent.
so, LSTM Meta Learner define $i_t$ as:

 $i_t=\sigma(W_I.[\bigtriangledown L_t,L_t,\theta_{t-1},i_{t-1}]+b_I)$
 
 i.e; a sigmoid function with a combination of current gradient, Loss, and previous learning rate $i_{t-1}$.
 
For $f_t$, it should be 1, but to avoid problems of shrinking gradients, and exploding gradients. It has been chosen as:

 $f_t=\sigma(W_F.[\bigtriangledown L_t,L_t,\theta_{t-1},f_{t-1}]+b_F)$
 
i.e; sigmoid function with a combination of current gradient, Loss, and forget gate.

\subsubsection{Data Pre-processing}
In Typical Machine Learning setting, give a data-set D, we divide into 2 parts: training set, and a test set. Whereas in Meta-Learning Setting, we have Meta-sets say $D$ , consists of multiple sets of different task examples. For each D $\in D$  consists of $D_{train}$ and $D_{test}$ , so for K-Shot Learning each $D_{train}$ consists of $K \times N$ examples, where $N$ is the number of classes.
In Meta-Learning, since beginning data is divided into 3 parts: $D_{meta-train}, D_{meta-validation},$ and  $D_{meta-test}$
Here the objective is to use $D_{meta-train}$  is for training a learning procedure that can take as input one of its training sets $D_{train}$  and produce a classifier (learner) that achieves high average classification performance.
\subsubsection{Model Training}
Following the concept of Matching Networks, This Optimization Algorithm also tried to match training conditions to test time conditions, as they have been proven to perform really well at a task. To make similar conditions,  when considering each dataset D , it uses loss $L_{test}$  produced by D's test set $D_{test}$ , for step by step knowledge refer to Algorithm mentioned in paper \cite{ravi2016optimization}.

\begin{figure}[t]
\begin{center}
   \includegraphics[width=1.0\linewidth]{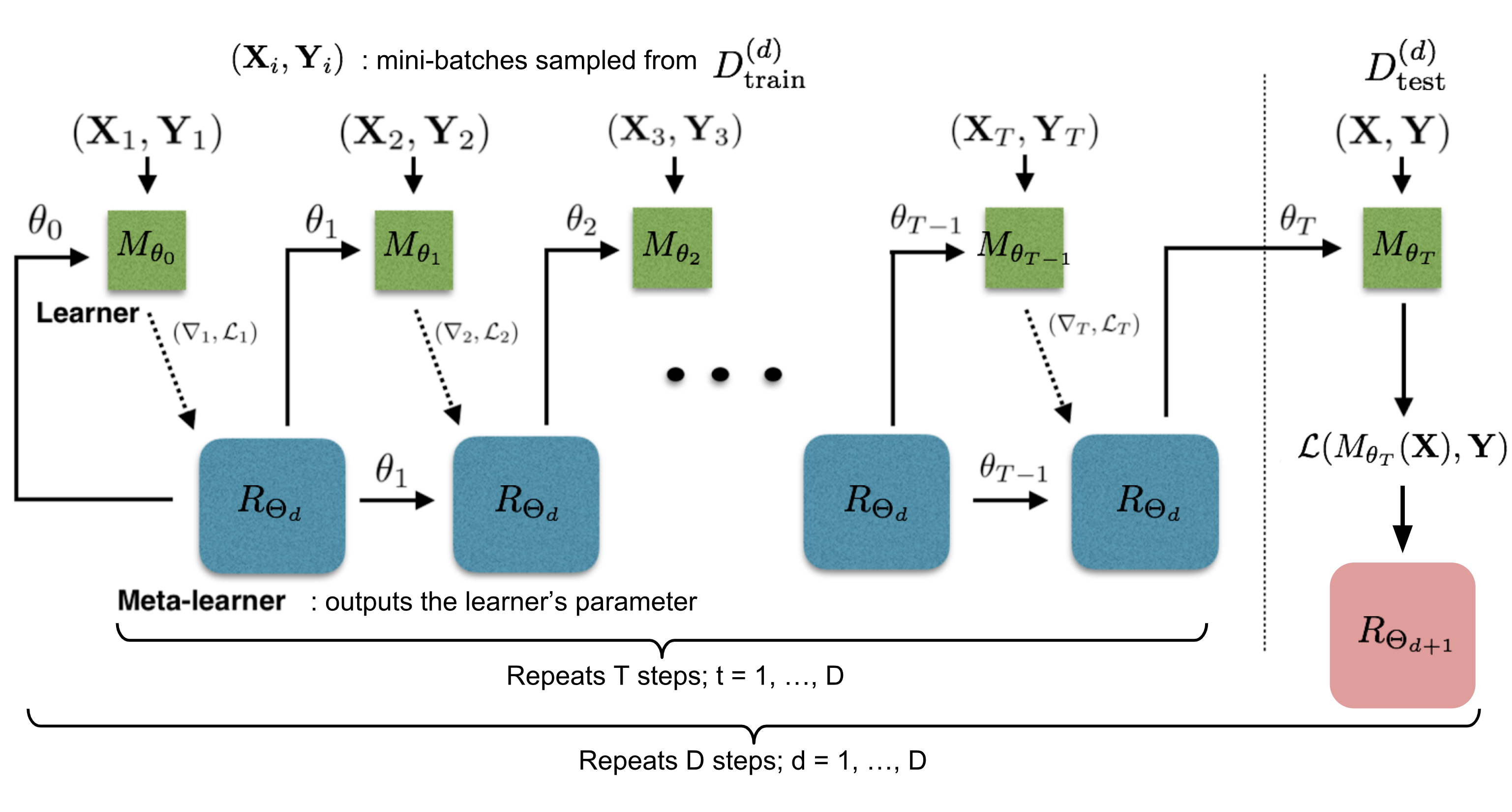}
\end{center}
   \caption{The architecture of LSTM Meta Networks \cite{ravi2016optimization} consists of 2 parts: Learner, and Meta-Learner. Learner trains for T-steps and sends its updated parameters to Meta-Learner, then Meta-Learner evaluates updates, and optimize its parameters. This whole process runs for D steps.}
\label{fig:11}
\end{figure}

So, while iterating for T steps overall D's training data, LSTM meta-learner receives some parameter updates from learner. After T-steps, final parameters are then used to evaluate Test set, and make updates on meta-learner. To understand, a pictorial representation of architecture, refer architecture in \ref{fig:11}

\section{Other Approaches for few-shot learning}
As we know few-shot Learning is a sub-field of Machine Learning, there are different relevant solutions which are very close to few-shot Learning approach, but yet different in their solution approach. Such problems can be solved by using few-shot Learning algorithms as well. Let's go through each of such fields of Machine Learning, and observe how they are close to few-shot Learning Problem:
\begin{enumerate}
    \item Semi-Supervised Learning
    \item Imbalanced Learning
    \item Transfer Learning
\end{enumerate}

\begin{figure}[t]
\begin{center}
   \includegraphics[width=0.7\linewidth]{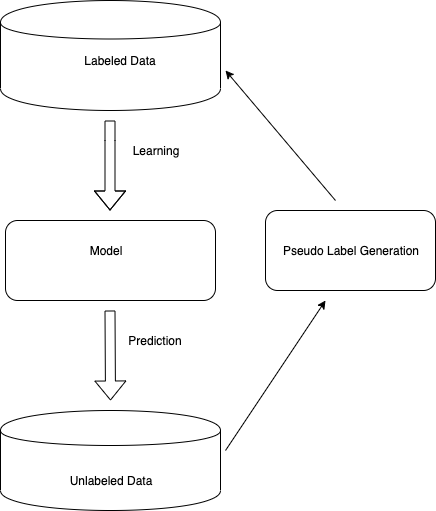}
\end{center}
   \caption{Semi Supervised Learning}
\label{fig:long}
\label{fig:onecol}
\end{figure}

\subsection{Semi-Supervised Learning}
Semi-supervised learning\cite{zhu2009introduction} is a machine learning technique that involves training a model on both labeled and unlabeled data. In traditional supervised learning, models are trained only on labeled data, which is expensive and time-consuming to obtain. However, with semi-supervised learning, models can leverage the vast amount of unlabeled data available to improve performance on labeled data.

Suppose we have 10,000 data points, out of which only 20,000 are labeled, and the rest 80,000 are unlabeled. In such cases, we can take the help of semi-supervised learning to train a more accurate model. Semi-supervised learning \cite{zhu2009introduction} goes through a pseudo-labeling technique to increase the training set. In this technique, we train a model using the 20,000 labeled dataset and use it on an equally sized test dataset to create pseudo-labels for the unlabeled data points. After obtaining pseudo-labels, we concatenate the real labels with the pseudo-labels and real features with the pseudo-features. We then train a new model on the concatenated dataset, which has been shown to be more accurate than the initial model. We repeat this process until optimal accuracy is achieved.

Semi-supervised learning works well in scenarios where labeled data is scarce, and there is a vast amount of unlabeled data available. By leveraging the unlabeled data, models can gain a better understanding of the population structure in general, leading to improved performance on the labeled data. Semi-supervised learning has been applied in various fields such as computer vision, natural language processing, and speech recognition.

However, it is important to note that semi-supervised learning is not a panacea, and it may not always lead to improved performance. The success of semi-supervised learning depends on various factors such as the quality of the unlabeled data, the complexity of the task, and the type of model used. Additionally, the pseudo-labeling technique may introduce errors if the model makes incorrect predictions on the unlabeled data. Therefore, it is crucial to validate the accuracy of the pseudo-labels before using them to train the model.

\subsection{Imbalanced Learning}
In Imbalanced Learning \cite{haixiang2017learning} scenario, we have an imbalanced data-set i.e; we have more samples from one class, whereas very few examples of other categories, this is also popularly known as skewed distribution data-set. Let’s take a look at some popular methods for dealing with a skewed data-set.
\begin{enumerate}
    \item Choice of Metric: There are various forms of metrics\cite{jadon2022comprehensive} that can help in assessing the accuracy of a model, such as Confusion Matrix, Precision, Recall, and F1-Score.
    \item Choice of Machine Learning Algorithm: Parametric Algorithms learn their parameters through the data-set, so if the data-set is biased, it is most likely that the parametric model will also be biased. Non-parametric approaches such as K-Nearest Neighbor, Decision Trees etc. Ensembles such as Random Forest, Adaboost, XG-Boost are proven to be the best approaches when it comes to a biased data-set.
    \item Choice of Data Sampling Methods: Data Sampling can also be considered to ensure that data-set doesn't remain skewed.
\end{enumerate}

This is close to few-shot Learning, as the machine learning model we are expected to create should be able to learn distribution from a few examples.

\subsection{Transfer Learning}
Transfer Learning \cite{pan2009survey} is a machine learning technique that involves the application of knowledge gained from one task to another related task. It is based on the idea that the knowledge acquired by a model while solving one task can be leveraged to help improve the performance of a different task.

The primary motivation behind transfer learning is to reduce the need for large amounts of labeled data for a new task. In many real-world scenarios, it is expensive and time-consuming to collect large amounts of labeled data. However, with transfer learning, the knowledge gained from solving a related task can be used to train a model on the new task, reducing the amount of labeled data required.

There are three main types of transfer learning:
\begin{enumerate}
    \item \textbf{Inductive Transfer Learning:} In this approach, the model is first trained on a large, diverse dataset and then fine-tuned on the new task. The idea behind this approach is that the model has already learned features that are relevant to the new task, and fine-tuning helps to refine those features.
    
    \item \textbf{Transductive Transfer Learning:} This approach involves transferring knowledge between related tasks, where the goal is to learn the mapping between two different domains. For example, mapping between two different languages, or between image and text domains.

    \item \textbf{Unsupervised Transfer Learning:} This approach involves using the knowledge gained from unsupervised learning to improve the performance of a supervised learning task. In unsupervised learning, the model learns to represent data in a way that captures its underlying structure. This representation can then be used to improve the performance of a supervised learning task that uses the same data.
\end{enumerate}

Transfer learning has been applied in various fields such as computer vision, natural language processing, and speech recognition. For instance, in computer vision, transfer learning has been used to improve the performance of object recognition, object detection, and image classification. Similarly, in natural language processing\cite{patil2022auto}, transfer learning has been used to improve the performance of tasks such as sentiment analysis, named entity recognition, and machine translation.

In summary, Transfer Learning is a powerful technique that enables models to leverage knowledge gained from one task to improve performance on another related task. It is particularly useful in scenarios where labeled data is scarce, and it has been applied in various fields to improve the performance of machine learning models.

\section{Conclusion}
In order to develop a flawless few-shot learning method, it is essential to have access to impeccable information embeddings, efficient memory storage, and advanced optimization algorithms beyond the scope of gradient descent. Our research delves into various algorithms, including Matching Networks and Meta Networks, to explore their efficacy in achieving state-of-the-art accuracy in classification tasks. However, for more complex tasks such as object detection and image segmentation, our models are still facing difficulties.

Although there are various theoretical applications for few-shot learning, it has only recently gained momentum in real-world scenarios. Advancements in this field have been applied to diverse domains, such as writing SQL codes, enhancing deformed medical images, and verifying signatures. Although many other domains are still under research, major companies such as OpenAI, Google, Microsoft, and Amazon are investing significantly in AI research.

The successful implementation of few-shot learning can revolutionize the world by creating a human-like brain capable of detecting rare diseases and optimizing supply-chain models to tackle global food crises. The possibilities are limitless, and solving the challenges associated with few-shot learning can be a game-changer for humanity.

\small
\nocite{*} 
\bibliographystyle{ieeetr}
\bibliography{IEEEexample}

\begin{thebibliography}{10}

\bibitem{jadon2020hands}
S.~Jadon and A.~Garg, {\em Hands-On One-shot Learning with Python}.
\newblock Packt Publishing, 2020.

\bibitem{jadon2020comparative}
S.~Jadon, O.~P. Leary, I.~Pan, T.~J. Harder, D.~W. Wright, L.~H. Merck, and
  D.~L. Merck, ``A comparative study of 2d image segmentation algorithms for
  traumatic brain lesions using ct data from the protectiii multicenter
  clinical trial,'' in {\em Medical Imaging 2020: Imaging Informatics for
  Healthcare, Research, and Applications}, vol.~11318, p.~113180Q,
  International Society for Optics and Photonics, 2020.

\bibitem{jadon_2018}
S.~Jadon, ``Introduction to different activation functions for deep learning,''
  Mar 2018.

\bibitem{koch2015siamese}
G.~Koch, R.~Zemel, and R.~Salakhutdinov, ``Siamese neural networks for one-shot
  image recognition,'' in {\em ICML deep learning workshop}, vol.~2, 2015.

\bibitem{jadon2019improving}
S.~Jadon and A.~A. Srinivasan, ``Improving siamese networks for one shot
  learning using kernel based activation functions,'' {\em arXiv preprint
  arXiv:1910.09798}, 2019.

\bibitem{vargas2020one}
C.~Vargas, Q.~Zhang, and E.~Izquierdo, ``One shot logo recognition based on
  siamese neural networks,'' in {\em Proceedings of the 2020 International
  Conference on Multimedia Retrieval}, pp.~321--325, 2020.

\bibitem{vinyals2016matching}
O.~Vinyals, C.~Blundell, T.~Lillicrap, D.~Wierstra, {\em et~al.}, ``Matching
  networks for one shot learning,'' in {\em Advances in neural information
  processing systems}, pp.~3630--3638, 2016.

\bibitem{jadon2020ssm}
S.~Jadon, ``Ssm-net for plants disease identification in lowdata regime,'' {\em
  arXiv preprint arXiv:2005.13140}, 2020.

\bibitem{graves2014neural}
A.~Graves, G.~Wayne, and I.~Danihelka, ``Neural turing machines,'' {\em arXiv
  preprint arXiv:1410.5401}, 2014.

\bibitem{santoro2016one}
A.~Santoro, S.~Bartunov, M.~Botvinick, D.~Wierstra, and T.~Lillicrap,
  ``One-shot learning with memory-augmented neural networks,'' {\em arXiv
  preprint arXiv:1605.06065}, 2016.

\bibitem{collier2018implementing}
M.~Collier and J.~Beel, ``Implementing neural turing machines,'' in {\em
  International Conference on Artificial Neural Networks}, pp.~94--104,
  Springer, 2018.

\bibitem{munkhdalai2017meta}
T.~Munkhdalai and H.~Yu, ``Meta networks,'' {\em Proceedings of machine
  learning research}, vol.~70, p.~2554, 2017.

\bibitem{finn2017model}
C.~Finn, P.~Abbeel, and S.~Levine, ``Model-agnostic meta-learning for fast
  adaptation of deep networks,'' {\em arXiv preprint arXiv:1703.03400}, 2017.

\bibitem{ravi2016optimization}
S.~Ravi and H.~Larochelle, ``Optimization as a model for few-shot learning,''
  2016.

\bibitem{zhu2009introduction}
X.~Zhu and A.~B. Goldberg, ``Introduction to semi-supervised learning,'' {\em
  Synthesis lectures on artificial intelligence and machine learning}, vol.~3,
  no.~1, pp.~1--130, 2009.

\bibitem{haixiang2017learning}
G.~Haixiang, L.~Yijing, J.~Shang, G.~Mingyun, H.~Yuanyue, and G.~Bing,
  ``Learning from class-imbalanced data: Review of methods and applications,''
  {\em Expert Systems with Applications}, vol.~73, pp.~220--239, 2017.

\bibitem{jadon2022comprehensive}
A.~Jadon, A.~Patil, and S.~Jadon, ``A comprehensive survey of regression based
  loss functions for time series forecasting,'' {\em arXiv preprint
  arXiv:2211.02989}, 2022.

\bibitem{pan2009survey}
S.~J. Pan and Q.~Yang, ``A survey on transfer learning,'' {\em IEEE
  Transactions on knowledge and data engineering}, vol.~22, no.~10,
  pp.~1345--1359, 2009.

\bibitem{patil2022auto}
A.~Patil and A.~Jadon, ``Auto-labelling of bug report using natural language
  processing,'' {\em arXiv preprint arXiv:2212.06334}, 2022.

\bibitem{keren2018weakly}
G.~Keren, M.~Schmitt, T.~Kehrenberg, and B.~Schuller, ``Weakly supervised
  one-shot detection with attention siamese networks,'' {\em stat}, vol.~1050,
  p.~12, 2018.

\bibitem{9116199}
S.~{Mshir} and M.~{Kaya}, ``Signature recognition using machine learning,'' in
  {\em 2020 8th International Symposium on Digital Forensics and Security
  (ISDFS)}, pp.~1--4, 2020.

\bibitem{thathachar2020encoder}
J.~Thathachar, T.~Kornuta, and A.~S. Ozcan, ``Encoder-decoder memory-augmented
  neural network architectures,'' Mar.~19 2020.
\newblock US Patent App. 16/135,990.

\bibitem{collier2019memory}
M.~Collier and J.~Beel, ``Memory-augmented neural networks for machine
  translation,'' {\em arXiv preprint arXiv:1909.08314}, 2019.

\bibitem{shen2017meta}
F.~Shen, S.~Yan, and G.~Zeng, ``Meta networks for neural style transfer,'' {\em
  arXiv preprint arXiv:1709.04111}, 2017.

\bibitem{chen2019image}
Z.~Chen, Y.~Fu, Y.-X. Wang, L.~Ma, W.~Liu, and M.~Hebert, ``Image deformation
  meta-networks for one-shot learning,'' in {\em Proceedings of the IEEE
  Conference on Computer Vision and Pattern Recognition}, pp.~8680--8689, 2019.

\bibitem{liu2020does}
Z.~Liu, R.~Zhang, Y.~Song, and M.~Zhang, ``When does maml work the best? an
  empirical study on model-agnostic meta-learning in nlp applications,'' {\em
  arXiv preprint arXiv:2005.11700}, 2020.

\bibitem{behl2019alpha}
H.~S. Behl, A.~G. Baydin, and P.~H. Torr, ``Alpha maml: Adaptive model-agnostic
  meta-learning,'' {\em arXiv preprint arXiv:1905.07435}, 2019.

\bibitem{li2019finding}
H.~Li, D.~Eigen, S.~Dodge, M.~Zeiler, and X.~Wang, ``Finding task-relevant
  features for few-shot learning by category traversal,'' in {\em Proceedings
  of the IEEE Conference on Computer Vision and Pattern Recognition},
  pp.~1--10, 2019.

\bibitem{rebuffi2020semi}
S.-A. Rebuffi, S.~Ehrhardt, K.~Han, A.~Vedaldi, and A.~Zisserman,
  ``Semi-supervised learning with scarce annotations,'' in {\em Proceedings of
  the IEEE/CVF Conference on Computer Vision and Pattern Recognition
  Workshops}, pp.~762--763, 2020.

\bibitem{huang2016learning}
C.~Huang, Y.~Li, C.~C. Loy, and X.~Tang, ``Learning deep representation for
  imbalanced classification,'' in {\em Proceedings of the IEEE conference on
  computer vision and pattern recognition}, pp.~5375--5384, 2016.

\bibitem{fernandez2018learning}
A.~Fern{\'a}ndez, S.~Garc{\'\i}a, M.~Galar, R.~C. Prati, B.~Krawczyk, and
  F.~Herrera, {\em Learning from imbalanced data sets}.
\newblock Springer, 2018.

\bibitem{zhuang2020comprehensive}
F.~Zhuang, Z.~Qi, K.~Duan, D.~Xi, Y.~Zhu, H.~Zhu, H.~Xiong, and Q.~He, ``A
  comprehensive survey on transfer learning,'' {\em Proceedings of the IEEE},
  2020.

\end{thebibliography}

\end{document}